\documentclass[mlmain, twocolumn]{jmlr}

\usepackage{longtable}

\usepackage{booktabs}
\usepackage[load-configurations=version-1]{siunitx} 

\usepackage[utf8]{inputenc}

\usepackage{microtype}

\usepackage{booktabs}
\usepackage{multirow}
\usepackage{listings}
\usepackage{algorithm2e}
\RestyleAlgo{ruled}
\usepackage{textcomp}
\usepackage{float}
\usepackage{graphicx}
\usepackage{amsmath}
\usepackage{makecell}
\usepackage[title]{appendix}
\usepackage[export]{adjustbox}

\usepackage[load-configurations=version-1]{siunitx} 

\newcommand{\doctor}{medical professional}
\newcommand{\doctors}{medical professionals}

\DeclareMathOperator*{\argmax}{arg\,max}

\theorembodyfont{\upshape}
\theoremheaderfont{\scshape}
\theorempostheader{:}
\theoremsep{\newline}

\jmlrvolume{}
\firstpageno{1}

\title[Learning functional sections]{Learning functional sections in medical conversations: \\ iterative pseudo-labeling and human-in-the-loop approach}

\author{
  \Name{Mengqian Wang}\thanks{Work done during an internship at Curai Health}
  \Email{mengqian@email.unc.edu}\\
   \addr University of North Carolina at Chapel Hill \\
  \Name{Ilya Valmianski} \Email{ilya@curai.com} \\
  \Name{Xavier Amatriain}
  \Email{xavier@curai.com} \\
  \Name{Anitha Kannan}
  \Email{anitha@curai.com} \\
  \addr Curai Health
  \vspace{-2em}
  }
 
\begin{document}

\maketitle

\begin{abstract}
Medical conversations between patients and \doctors~have implicit functional sections, such as ``history taking", ``summarization'', ``education'',  and ``care plan.'' In this work, we are interested in learning to automatically extract these sections. A direct approach would require collecting large amounts of {\it expert annotations} for this task, which is inherently costly due to the contextual inter-and-intra variability between these sections. This paper presents an approach that tackles the problem of learning to classify medical dialogue into functional sections without requiring a large number of annotations. Our approach combines pseudo-labeling and human-in-the-loop. First, we bootstrap using weak supervision with pseudo-labeling to generate dialogue turn-level pseudo-labels and train a transformer-based model, which is then applied to individual sentences to create noisy sentence-level labels. Second, we iteratively refine sentence-level labels using a cluster-based human-in-the-loop approach. Each iteration requires only a {\it few dozen} annotator decisions. We evaluate the results on an expert-annotated dataset of 100 dialogues and find that while our models start with 69.5\% accuracy, we can iteratively improve it to  82.5\%. Code used to perform all experiments described in this paper can be found here: \url{https://github.com/curai/curai-research/tree/main/functional-sections}.

\end{abstract}


\vspace{-1em}
\section{Introduction}
\label{sec:intro}
Recent growth in telemedicine has led to a dramatic expansion in text-based chat communications between patients and \doctors~\citep{mckinsey-telehealth}. This creates new opportunities for improving \doctor~workflows through the introduction of natural language understanding (NLU) systems for providing real-time decision support and automating electronic health record (EHR) charting \citep{DreisbachCaitlin2019Asro, joshi2020dr, pmlr-v158-valmianski21a}. Auto-charting, in particular, benefits significantly from proper contextualization of the dialogue \citep{medfilter, Krishna2021GeneratingSN}. For example, the History of Present Illness (HPI) section of the progress note can be derived from the history-taking discussion in the dialogue, while the Care Plan section can be derived from the care plan discussion. 



\begin{figure*}[!ht]
    \centering
    \includegraphics[width=1\linewidth]{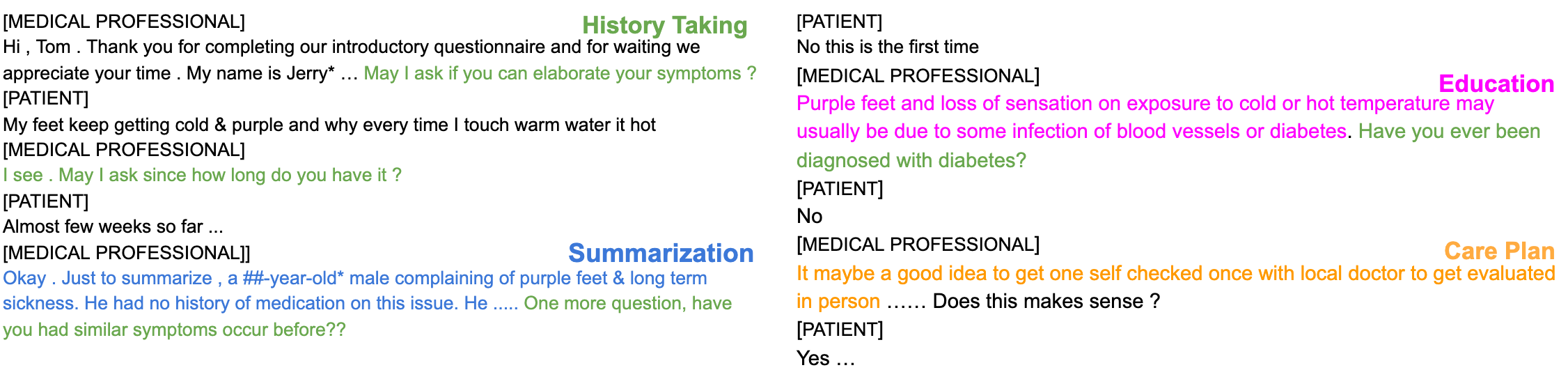}
         \vspace{-2em}
    \caption{A de-identified patient-\doctor~dialog color-coded with the predictions from our approach (best seen in color). Notice how different multiple labels may be present in the same turn of the conversation.}
    \label{fig:example-task}
    \vspace{-2.2em}
\end{figure*}


Similar to medical SOAP note \citep{soap-note} sections, we consider labeling each \doctor~written sentence into the following functional sections: 
\begin{enumerate}
    \vspace{-0.5em}
    \item \textit{History taking}: questions about the patient's current illness including symptoms, prior medical history, and medications they may be taking.
     \vspace{-0.8em}
    \item \textit{Summarization}: \doctor's confirmation of relevant patient symptomatology.
     \vspace{-0.8em}
    \item \textit{Education}: education of the patient about their medical issues.
     \vspace{-0.8em}
    \item \textit{Care plan}: suggestion on the course of action or treatments.
     \vspace{-0.8em}
    \item \textit{Other}: non-medical text.
\end{enumerate}

\autoref{fig:example-task} shows an abridged dialogue that is color-coded with appropriate section labels, as predicted by the model introduced in this paper. We make two observations. First,  within a single turn of the dialogue, multiple functional sections (or classes) can co-occur \textit{e.g.}  educating the patient that purple feet are a diabetes-related symptom while taking history on whether the patient has been previously diagnosed with diabetes. Second, dialogue sentences belonging to different functional sections may have high lexical overlap, \textit{e.g.} purple feet being discussed in the context of history taking, summarization, and education.



We formulate the problem of inferring conversation functional sections as sentence-level classification: {\it Given a \doctor-patient dialogue, how can we assign every sentence in every turn of the dialogue to the correct functional section?} Further, {\it How can we learn such a model when we can get only small amounts of human-generated labels?}

We tackle both these questions by leveraging two key insights:
\begin{enumerate}
\vspace{-0.5em}
    \item \emph{Dialogue turns carry more information than individual sentences and are thus easier to learn with weak supervision.} We use this insight to build a noisy turn-level labels dataset and train a language model to classify turn-level labels. We then apply the turn level model to label individual sentences within the turn, creating noisy sentence-level labels.
    \item \emph{Text embeddings learned from noisy labeled data are relevant even when the classifications are not reliable.} We use this insight to propose an iterative human-in-the-loop cluster-based pseudo-labeling strategy. Our proposed clustering strategy introduces variability in samples across iterations by enabling intermixing high-confidence predictions with low-confidence ones and choosing only class-specific `pure' clusters through a simple human-in-the-loop evaluation. 
\end{enumerate}

We evaluate the results on an expert-annotated dataset of 100 dialogues and find that although the initial pseudo-labels have an accuracy of 69.5\%, our iterative refinement approach can boost accuracy to 82.5\%. We also find that the latent space representations of each class become both more tightly clustered and more separable between different classes, which may imply higher generalizability \citep{Li2020OnTS}.

\vspace{-1em}
\section{Related Work}
\label{sec:related_work}
\paragraph*{Semantic structure understanding:}
The importance of identifying and assigning labels to functionally coherent units is well-understood. As an example, in legal document understanding, \cite{saravanan-etal-2008-automatic,Malik2021SemanticSO} show that it's easier for downstream tasks if documents are segmented into coherent units such as facts, arguments, statutes, {\it etc}. In conversational dialogues, the problem of utterance-level intent classification to detect discourse boundaries is well studied \citep{liu17, raheja19, Qu19, Joty14, Takanobu18}. These intents are  broad ({\it e.g.}``original question" and ``repeat question"  \citet{Qu19}) and identified at turn-level.


We are interested in classifying dialogue turns and also each sentence within a turn into functional sections (history taking, summary, education, care plan, other) that can loosely serve as intents. These sections interleave ({\it e.g.} history taking and education) within  a single dialogue turn making the task challenging. Previous works assume access to manually labeled data. In this paper, we bootstrap data using a weak pseudo-labeler and then iteratively refine it with training text-classification models, clustering their embeddings, and relabeling entire clusters using a human-in-the-loop. 

\paragraph*{Active learning:}  This approach focuses on starting with a small labeled dataset and iteratively retraining models with an updated labeled dataset (see references in survey papers \cite{settles2009active} and \cite{Ren2020}). Each update to the labeled training set involves getting manual labels for a small (often only one) number of  most informative examples - examples of which the model at the previous iteration is most uncertain. 

In contrast, our human-in-the-loop approach aims to change labels in a much larger number of examples in each turn. We cluster the embeddings of the examples based on the current model, get cluster-level annotation from the human annotators, and impute that label to all the examples of the cluster. Of related is the work of  \cite{mottaghi2020} that uses clustering within the active learning framework, but the technique was used to only identify previously unseen classes and to obtain a small number of informative examples within each cluster to increase coverage.

 \paragraph*{Pseudo-labeling:} Another approach to impute labels to a large number of unlabeled examples is to  \citep{mindermann2021, du2020, chen2020} use a trained model's prediction to self-label (self-training or pseudo-labeling \citet{Lee13}). 
 
\citet{du2020} shows that self-training with pseudo-labeling can improve performance on text classification benchmarks without the need for in-domain unlabeled data.  While being general and domain-agnostic, pseudo-labeling approaches can under-perform if the generated labels are noisy (e.g., high variance model in the previous iteration of training) and hence adversely affect performance (c.f. \citep{Oliver2018,Rizve2021, Nair2021} and references therein). In this paper, we combine pseudo-labeling followed by independent clustering of the pseudo-labeled-class specific data points. Human experts then annotate samples from each cluster to either relabel the entire cluster or remove it from the next  training iteration (because it contains sentences from multiple functional sections).

\section{Approach}
\label{sec:approach}

In this section, we present a general description of our approach. We describe the specifics of applying this approach to medical conversations in \S~\ref{sec:exp-details}.

\vspace{-0.5em}
\begin{table}[H]
\begin{tabular}{p{0.2\linewidth} p{0.7\linewidth}}
\textbf{Symbol} & \textbf{Description} \\ 
$D$ & Dialogue \\
$T_{i}$ & i'th turn of the dialogue \\
$S_{ij}$ & j'th sentence in the i'th turn of the dialogue \\
$\mathcal{L}$ & Universe of labels \\
$L^{\textnormal{turn}}_{i}$ & Turn level functional section label of the i'th turn \\
$L^k_{ij}$ & k'th iteration sentence level functional section label of i'th turn and j'th sentence \\
$M_{turn}$ & Text classification model trained on turn level functional section labels \\
$M_{\textnormal{sent}}^k$ & k'th iteration text classification model trained on sentence level functional section labels \\
$\hat{L}^{k}_{ij}$ & Estimated labels for $S_{ij}$ produced by $M_{\textnormal{sent}}^k$\\
$E^k_{ij}$ & Embedding of the $S_{ij}$ by $M_{\textnormal{sent}}^k$ \\
$\textnormal{Clst}$ & Clustering algorithm applied independently to embeddings $E$ for each functional section in $\mathcal{L}$ \\
$C^k_{ij}$ & Cluster assigned to $S_{ij}$ by $\textnormal{Clst}$ using embeddings and labels produced by $M_{\textnormal{sent}}^k$. \\
$H$ & Human annotator that reviews a set of sentences and assigns the set one of the labels in $\mathcal{L}$ or marks them as ``Mixed'' \\
$L_n^{\textnormal{clst}}$ & A label applied to all sentences of cluster $n$ ({\it e.g. $C^k_{ij}=n$})

\end{tabular}
\vspace{-2em}
\caption{Notation used in this paper.}
\label{table:notation}
\end{table}

\autoref{fig:approach-overview} presents a schematic overview of our approach. It consists of two parts. First, in turn-to-sentence label bootstrapping, we pseudo-label turn-level labels \footnote{We use labels and functional sections interchangeably based on the context} which we use to train a text classification model. We then apply this model to sentences to create noisy sentence-level labels (\S~\ref{sec:approach-turn-to-sent}). Second, we iterate on sentence-level labels by training a text classification model, clustering the sentence-level embeddings, and then using a human-in-the-loop to classify the sentences of each cluster (\S~\ref{sec:iter-sent-refine}). The notation used in this paper is described in \autoref{table:notation}.


\subsection{Turn-to-sentence label bootstrapping}
\label{sec:approach-turn-to-sent}

 \begin{figure}[!ht]
    \centering
    \includegraphics[width=1\linewidth]{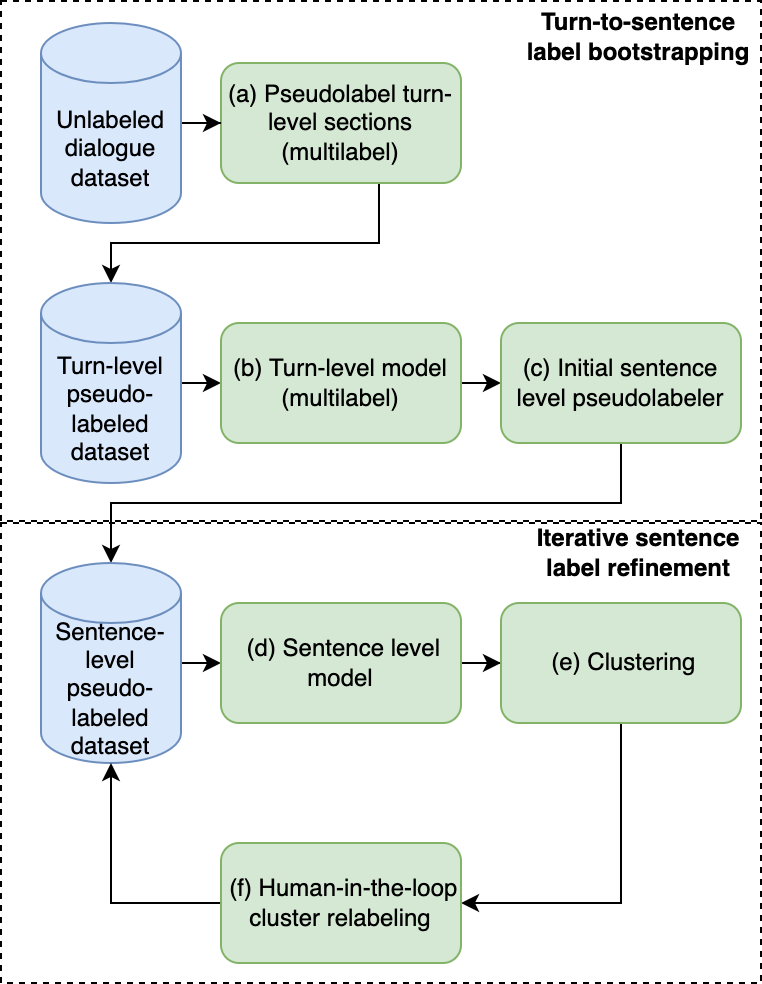}
    \caption{Schematic of the approach. The output of this approach is both the labeled dataset (after step f), and a classification model (step d).}
    \label{fig:approach-overview}
\end{figure}
In this step, we  train a turn-based model (\autoref{fig:approach-overview}b) to serve as the noisy pseudo-labeling for sentence-level labeling. We use a set of weak labelers to generate a turn level multilabel dataset for this task (\autoref{fig:approach-overview}a) of the form: $L^{\textnormal{turn}}_i = \cup_j  L_{ij}$. 




 $L^{\textnormal{turn}}$ are used to train a turn-level multilabel model $M_{\textnormal{turn}}$ (\autoref{fig:approach-overview}b). $M_{\textnormal{turn}}$ is then used to generate sentence-level labels by applying directly on sentences instead of entire turns, $L^0_{ij} \leftarrow M_{\textnormal{turn}}(S_{ij}, D)$ (\autoref{fig:approach-overview}c). Note that in \S~\ref{sec:application-turn} we discuss how, in our application, we still find it useful to apply (simple) rules on top of trained model output.
\subsection{Iterative sentence label refinement}
\label{sec:iter-sent-refine}
The iterative refinement starts with training a sentence-level text classification model (\autoref{fig:approach-overview}d), which is then used to produce both estimated labels and embedding $(\hat{L}^{k}_{ij}, E^k_{ij})$. These embeddings are clustered independently for each functional section label (\autoref{fig:approach-overview}e), and then each cluster, as a whole, is relabeled by a human annotator (\autoref{fig:approach-overview}f). Examples from clusters that contain sentences belonging to different functional sections (as judged by a human annotator and marked as ``Mixed'') are not used in the next iteration of retraining.  See \autoref{alg:refinement} for details.

\begin{figure}[ht]
\centering
\begin{minipage}{0.9\linewidth}
\begin{algorithm}[H]

	\setcounter{AlgoLine}{0}
	\LinesNumbered 
	\setcounter{AlgoLine}{0}
	\BlankLine
	\LinesNumbered 
	\SetKwInOut{Input}{Input}
    \SetKwInOut{Output}{Output}
	
	\Input{Dialogue dataset $\mathcal{D}$ \\
	       Current iter. model $M^k_{\textnormal{sent}}$ \\
	       Current iter. pseudo-labels $L^k_{ij}$ \\
	       Clustering algorithm $\textnormal{Clst}$ \\
	       Cluster annotator $H$}
	       
	\Output{$\{L^{k+1}_{ij}\}$}
	\BlankLine	
	
    $\{(\hat{L}^{k}_{ij}, E^k_{ij})\} \leftarrow \\ \{\forall D \in \mathcal{D}, \forall S_{ij} \in D, M^k_{\textnormal{sent}}(S_{ij}, D)\}$ \\
    
    $\{C_{ij}^k\} \leftarrow \textnormal{Clst}(\{(\hat{L}^{k}_{ij}, E^k_{ij})\})$ \\

    $\{S_{ij}\}_n \leftarrow \textnormal{Sample}(\{S_{ij} : C^k_{ij}=n\})$ \\

    $L^{\textnormal{clst}}_n = H(\{S_{ij}\}_n)$ \\
    
    $ L^{k+1}_{ij} \leftarrow L^{clst}_n : n=C^k_{ij} $

	\Return $\{L^{k+1}_{ij}: L^{k+1}_{ij} \ne \textnormal{Mixed}\}$    
	\vspace{5pt}
		\caption{Pseudocode for iterative cluster refinement of sentence level models. Sample function draws a small number  of examples ({\it e.g.} 10) from a set. ``Mixed'' represents that the set of sentences has sentences that pertain to several different functional sections.}
    \label{alg:refinement}

\end{algorithm}
\end{minipage}
\end{figure}


\section{Experimental details}
\label{sec:exp-details}

\subsection{Dataset}
\label{sec:dataset}
We use a dataset with 60,000 \doctor-patient encounters containing over 900,000 dialogue turns and  3,000,000 sentences collected on Curai Health virtual primary care platform. To construct a test set, we randomly sampled 100 encounters (not used for training or validation) for which we procured human labels for all \doctor~written sentences (3,102 sentences). In the human-labeled dataset, the distribution of sections on the sentence and turn levels are respectively: summarization: 3.6\%, 2.6\%; history taking: 26.5\%, 31.7\%; education: 5.3\%, 8.4\%; care plan: 4.1\%, 7.9\%; other: 60.3\%, 49.3\%.  We do {\bf not} have any additional labels for these encounters.


\subsection{Turn-to-sentence label bootstrapping}
\label{sec:application-turn}
As described in \S~\ref{sec:approach-turn-to-sent}, we first generate a dataset using {\it ad hoc} methods to train a turn level multilabel classification model. We then use this model to pseudo-label individual sentences. 



\paragraph*{Unsupervised clustering and human annotation of clusters.} We embed dialogue turns into fixed-sized representations by mean-pooling the final layer of the off-the-shelf DeCLUTR\footnote{We also tried  BioBERT \citep{biobert}, Mirror-BERT \citep{mirrorbert}, and Sentence BERT \citep{sbert}, but found that DeCLUTR produces representations that cluster with high label-purity}  \citep{declutr} sentence encoder. Following \citet{umap-for-clustering}, we project the 768D original embedding space to 250D via PCA and then project via UMAP \citep{McInnes2018} to 50D. We then cluster these 50D representations using the k-means++ algorithm \citep{kmeans} and determine the number of clusters using the elbow method \citep{Thorndike53whobelongs} (in our dataset, this number was 10). Human annotators manually label the resulting clusters by examining a small number ($\sim$ 10) of sentences in each cluster \autoref{fig:cluster}.
\paragraph*{Human annotation of a cluster-derived set of examples.} Because the unsupervised clustering did not produce good clusters containing only education or care plan turns, we procured human labels for 5000 turns from a mixed cluster containing education and care plan turns. 
\paragraph*{String-based rules.}  We identify turns with summarization sentences by string matching one of ['summar', 'sum up'].


\paragraph*{Turn-level model to generate sentence pseudo-labels.} We construct the dataset for the turn-level model by assigning the same label as the cluster after removing all mixed clusters. We then train $M_{\textnormal{turn}}$, a  multi-label classifier on top of DeCLUTR using this turn-level labeled set. The classification head consists of a single feed-forward layer with sigmoidal activation for each label.

To create the initial sentence level labels, we apply the turn-level model on each sentence and assign labels according to \autoref{algo:turn-to-sent} in the Appendix.

\begin{table}
\begin{adjustbox}{width=\columnwidth,center}
\begin{tabular}{llll}
\toprule
 & \multicolumn{3}{c}{\textbf {Round transition}} \\
  & \textbf{1\textrightarrow{}2} & \textbf{ 2\textrightarrow{}3} & \textbf{ 3\textrightarrow{}4} \\
 \midrule
\textbf{History taking} & - & 1/10\textrightarrow{}M & 3/10\textrightarrow{}M \vspace{5pt}\\ 
\textbf{Summarization} & 2/10\textrightarrow{}M & 1/10\textrightarrow{}M  & 3/10\textrightarrow{}M \\
&&1/10\textrightarrow{}O&\vspace{5pt}\\
\textbf{Education} & 1/10\textrightarrow{}M & 6/10\textrightarrow{}M &3/10\textrightarrow{}M \\
&&&1/10\textrightarrow{}O\vspace{5pt}\\
\textbf{Care plan} & 1/10\textrightarrow{}M & 3/10\textrightarrow{}M & 6/10\textrightarrow{}M \vspace{5pt}\\
\textbf{Other (O)} & 7/15\textrightarrow{}M & 3/10\textrightarrow{}M & 6/10\textrightarrow{}M \\ \bottomrule
\end{tabular}
\end{adjustbox}
\caption{Cluster relabeling between rounds.  Elements of the table correspond to how many clusters of a given semantic class were re-labeled as (O)ther or (M)ixed. }
\label{tab:iteration-details}
\end{table}

\subsection{Iterative sentence label refinement}
\label{sec:application-sent}

\paragraph*{Sentence-level model.} The input to this model is the dialogue turn that contains the target sentence. We mark the target sentence with tokens \textlangle{}START\textrangle{} and \textlangle{}END\textrangle{}. The model itself consists of a transformer language model DeCLUTR sentence encoder, with a classification head consisting of a single feed-forward layer with a softmax activation.
 
\paragraph*{Clustering sentence-level model.} To cluster sentence-level embeddings, we use a similar approach to the one described in turn-level clustering (\S~\ref{sec:application-turn}). The only difference is that we use the predicted labels to constrain that the kmeans++ algorithm is independently applied to examples corresponding to each predicted label.  As an example, \autoref{fig:cluster} in the Appendix shows the visualization of clusters predicted to be part of  "Summarization."  Each cluster is manually assigned its label (often simply staying with the original predicted label) by examining about ten data points (sentences). 

\paragraph*{Details of relabeling between rounds.} \autoref{tab:iteration-details} shows the number of clusters relabeled and the new label assigned. We can see that most relabeling was moving clusters to the ``Mixed" label, thereby ensuring that we improve the `purity' of the pseudo-labels.   Examples with the ``Mixed'' label are not  used for the subsequent round of model training.  However, they would still be used for subsequent clustering and relabeling. This strategy of relabeling also helps to mix high-confidence predictions with low-confidence ones, as long as they are close in the embedding representations.

\subsection{Implementation details}
All models discussed are trained in Pytorch 1.10.2+cu102 with the language models implemented using HuggingFace Transformers library \citep{Wolf2019HuggingFacesTS}. The weights for the DeCLUTR models were using the \textsc{johngiorgi/declutr-base} checkpoint. For training, we used the Adam optimizer with learning rate $2e^{-5}$ and a scheduler with warm-up steps of $\text{total training steps}/5$. We set the batch size as 12. PCA and kmeans were implemented using scikit-learn 0.24.2 package, while UMAP used the umap-learn 0.5.1 package.

\vspace{-0.5em}
\section{Results}

 \begin{figure*}[!htb]
    \centering
    \includegraphics[width=0.9\linewidth]{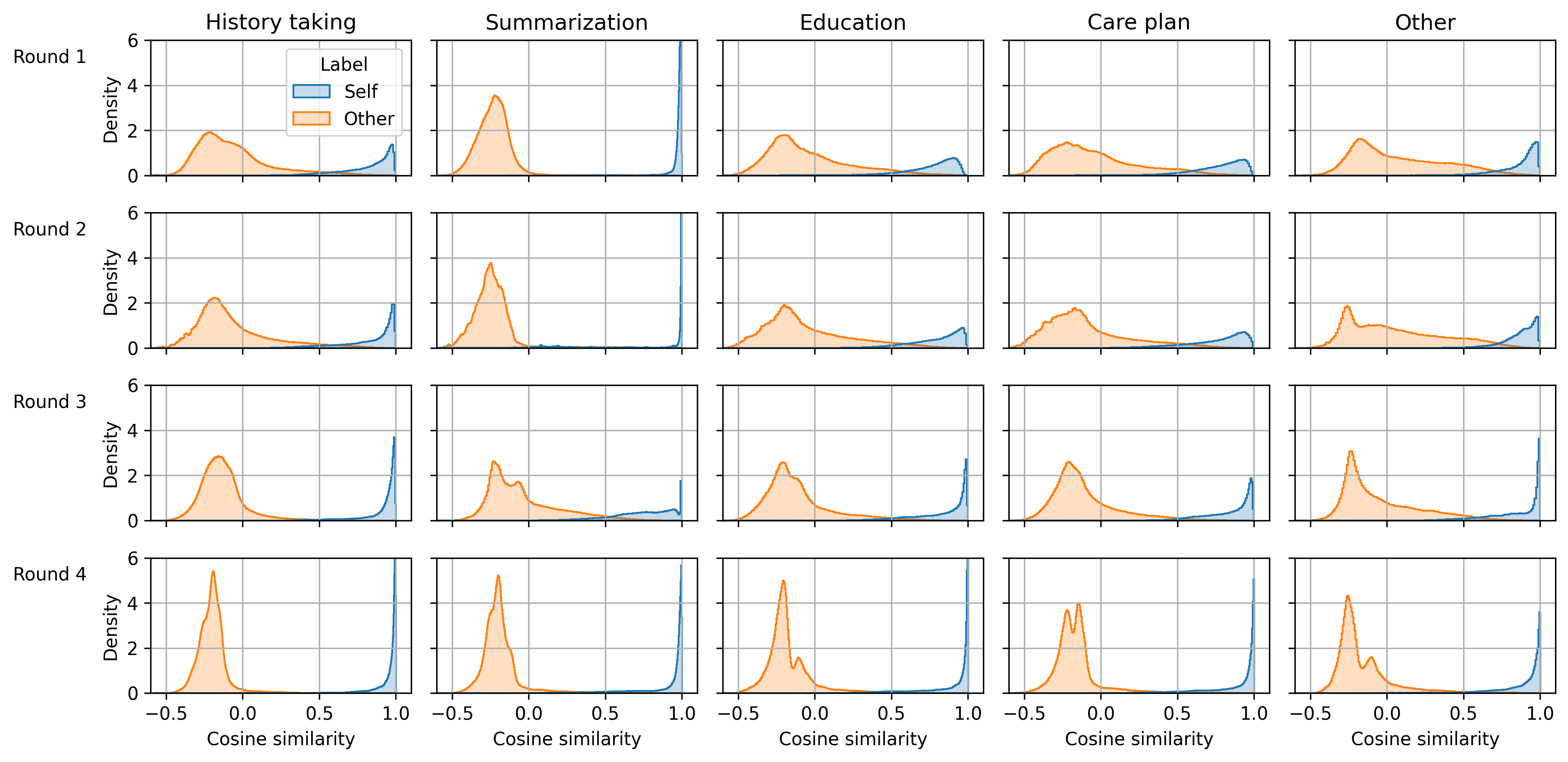}
    \caption{Cosine similarity of same- and different- class pairs for each class}
    \label{fig:cosine-sim}
\end{figure*}

\begin{figure*}[hbt]
    \centering
    \includegraphics[width=0.9\linewidth]{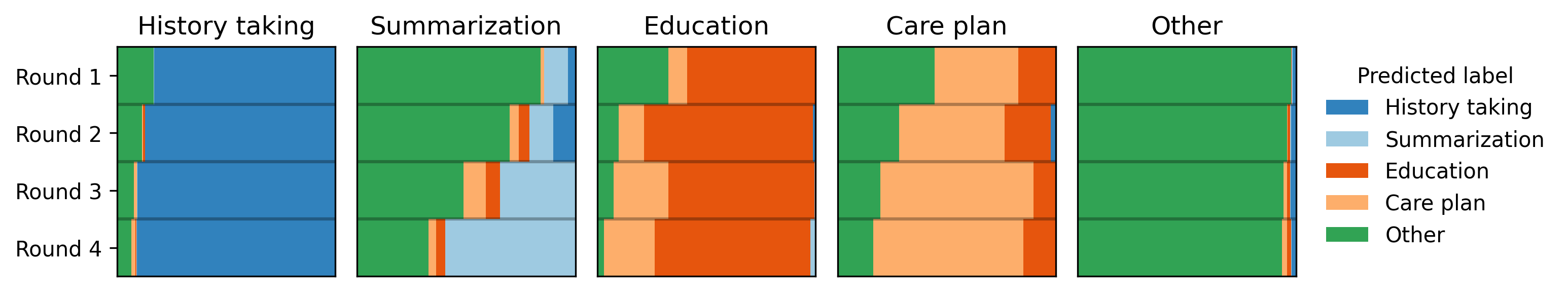}
    \caption{Improvement in label purity with each round (\S~\ref{sec:main_results}). Columns represent human-assigned true labels. Each row represents the proportion of the predicted labels as represented by the proportion of the color. The larger the proportion of the color corresponding to the column name, the better is the model improvement.}
    \label{fig:proportion}
    \vspace{-2em}
\end{figure*}

\begin{table*}[ht!]
\centering

\begin{tabular}{ccccc}
\hline
 & \multicolumn{4}{c}{\textbf{F1 score}} \\
\textbf{Class} & \textbf{Round 1} & \textbf{Round 2} & \textbf{Round 3} & \textbf{Round 4}\\
\hline
Summarization & 0.18$\pm$0.00 & 0.19$\pm$0.11 & 0.47$\pm$0.06 & \textbf{0.65}$\pm$\textbf{0.02}\\
History Taking & 0.89$\pm$0.01 & 0.9$\pm$0.00 & 0.92$\pm$0.00 & \textbf{0.93}$\pm$\textbf{0.01}\\
Education & \textbf{0.70}$\pm$\textbf{0.01} & 0.69$\pm$0.02 & 0.69$\pm$0.02 & 0.65$\pm$0.02\\ 
Care Plan & 0.55$\pm$0.02 & 0.56$\pm$0.03 & \textbf{0.57}$\pm$\textbf{0.01} & 0.55$\pm$0.02\\ 
Other & 0.90$\pm$0.00 & 0.92$\pm$0.00 & \textbf{0.93}$\pm$\textbf{0.00} & \textbf{0.93}$\pm$\textbf{0.00}\\
\Xhline{4\arrayrulewidth}
\textbf{Multi-class Accuracy*} & 69.5\%$\pm$0.00 & 74.1\%$\pm$0.00 & 80.4\%$\pm$0.01 & \textbf{82.5\%}$\pm$\textbf{0.01}\\
\hline
\multicolumn{5}{l}{* Accuracy is on the four functional classes only}\\
\end{tabular}

\caption{Sentence-level model performance: F1 scores and accuracy after each round of iterative training.  Standard deviations by retraining models with different seeds.}
\label{tab:sent-level}

\begin{tabular}{cccccc}
\hline
 & \multicolumn{5}{c}{\textbf{F1 score}} \\
\textbf{Class} & \textbf{Turn-level} & \textbf{Round 1} & \textbf{Round 2} & \textbf{Round 3} & \textbf{Round 4}\\
\hline
Summarization & 0.22$\pm$0.00 & 0.22$\pm$0.00 & 0.25$\pm$0.03 & \textbf{0.69}$\pm$\textbf{0.04} & 0.66$\pm$0.04\\
History Taking & 0.37$\pm$0.00 & 0.84$\pm$0.02 & 0.83$\pm$0.01 & 0.86$\pm$0.01 & \textbf{0.87}$\pm$\textbf{0.01}\\
Education & 0.61$\pm$0.01 & \textbf{0.77}$\pm$\textbf{0.02} & 0.69$\pm$0.05 & 0.73$\pm$0.02 & 0.65$\pm$0.04 \\ 
Care Plan & 0.31$\pm$0.04 & 0.55$\pm$0.02 & 0.55$\pm$0.03 & \textbf{0.57}$\pm$\textbf{0.01} & 0.51$\pm$0.02\\ 
Other & 0.75$\pm$0.00 & 0.89$\pm$0.01 & 0.93$\pm$0.00 & \textbf{0.95}$\pm$\textbf{0.01} & \textbf{0.95}$\pm$\textbf{0.00}\\
\Xhline{4\arrayrulewidth}
\textbf{Binary Accuracy} & 84.7\%$\pm$0.00 & \textbf{95.6}\%$\pm$0.00 & 95.2\%$\pm$0.01 & \textbf{95.6}\%$\pm$\textbf{0.00} & 94.9\%$\pm$0.00 \\
\hline
\multicolumn{6}{l}{* Accuracy is on the four functional classes only}\\
\end{tabular}

\caption{Turn-based inference improved with sentence-level model (~\S~\ref{sec:turnresults}). The column ``Turn-level" is the initial turn-level model from which sentence level model was bootstrapped. Columns Round 1--4 show the F1-score when we pool sentence-level predictions to produce turn level labels. The standard deviations are derived by retraining models with different seeds.}
\label{tab:turn-level}
\vspace{-2em}
\end{table*}

\label{sec:results}
\subsection{Main result: Sentence-level model performance}
\label{sec:main_results}
\autoref{tab:sent-level} provides our main results, comparing F1 and accuracy scores from each training round of the sentence-level model. The overall performance increased from accuracy of 69.5\% to 82.5\%. The ``Summarization" class has the most improvement (F1 score from 0.18 to 0.65). This three-fold improvement of the F1 score shows that our iterative approach can improve labeling quality (and hence the model) even when the initial labels are noisy. The sentences in this class are hard to identify solely from the turn-level-model-based pseudo-labeling. The pseudo-labeler successfully labels the sentences that contain ``to summarize" but fails on \textit{e.g.} "he experiences no pain."  However, our iterative clustering-based labeling introduces less-confident predictions that are semantically similar to the more confident ones to improve the overall identifiability.  

\autoref{fig:proportion} provides a graphical representation of the errors the model makes. Each column represents the human-assigned true label, and each row represents the proportion of the predicted labels in each true label for each training round. The two classes that see the F1 score uplift, ``History taking" and ``Summarization",  start with a significant confusion with the ``Other" class, which gradually decreases. Even though the additional iterations did not improve the ``Care plan'' and ``Education'' classes,  their overall confusion changed between rounds. Initially, both ``Education" and ``Care plan" were confused with the ``Other" class, while in later rounds, they were confused with each other. We expect this inter-class confusion as they can be hard to differentiate even for human annotators, \textit{e.g.} ``It is recommended that a person having a fever should drink more water.'' could be annotated as either ``Education" or ``Care plan", depending on the context.

\autoref{fig:cosine-sim} sheds light on another perspective on the change in the quality of the embeddings of the sentence-level models. Here, at every round, we randomly sampled 1,000 examples for each predicted class and used their embeddings to compute the distribution of cosine similarities between pairs of the same class (``self") and pairs of different classes (``other"). The distributions are always bimodal, but the full width at half max of the peaks decreases. Even for classes where the F1 metrics did not improve, there is an increase in the `peakiness" of the two distributions, making them more separable. This is the separation between positive and negative contrastive learning examples, where recent literature on sentence embeddings \citep{Li2020OnTS, mirrorbert} suggests that the increased separation corresponds to better generalization performance. 

\subsection{Can we obtain a better turn-level inference using the sentence-level model?}
\label{sec:turnresults}
In the previous experiments, we evaluated the output of the sentence-level model for each sentence in the input. Here, we investigate if training models at the sentence level also improve turn-level performance. For this, we  max pool the predictions of all the sentences in a turn.  For comparison, we use the initial turn level model  (\S~\ref{sec:approach-turn-to-sent}) as the baseline. 



\autoref{tab:turn-level} shows the F1 and accuracy scores of the sentence-aggregated turn-level predictions. Like the sentence-level models, we see the most marked improvement in the ``Summarization" class. Note how the Round 1 sentence-level model outperforms the turn-level model even though the turn-level model is used to generate the sentence-level pseudo-labels at the beginning with no human relabeling. This suggests that the sentence-level model can learn better semantics that the turn-level model. 

Overall, the improvement from the later rounds is less pronounced at the turn level. While sentence-level evaluation benefits from multiple rounds of disentangling the class confusion between sentences within a turn, this is less of a concern for turn-level evaluation. This is also evidenced by overall higher F1 scores when compared to evaluation at the sentence level in \autoref{tab:sent-level}. The improved performance and decreased effect from additional iterations is likely because the sentences that are more difficult to classify into a particular class tend to appear in mixed class turns and therefore doing well on these sentences does not improve turn-level metrics. 





\section{Discussion}
\label{sec:discussion}

We proposed a method for automatically inferring functional sections of a patient-\doctor~dialogue with minimal human supervisory data. While we focused on the four dominant medically relevant functional sections,  ``History taking", ``Summarization," ``Education," and ``Care plan" along with a background (``Other'') class, the approach can be easily extended to additional classes.

Starting with very little annotated data, we build a highly accurate model using a human-in-the-loop cluster-based pseudo-labeling strategy. We show that the approach increases embedding anisotropy, effectively increasing the contrast between labels. We think this is because our approach intermixes high and low-confidence predictions which are then relabeled on a per-cluster basis through a simple human-in-the-loop evaluation.  This makes our label-refinement strategy potentially useful for other applications, where the starting pseudo-labels are noisy or insufficient to capture the data variability, and getting additional human labels is expensive.

\paragraph*{Ethics} This work was done as part of a quality improvement activity as defined in 45CFR §46.104(d)(4)(iii) -- secondary research for which consent is not required for the purposes of “health care operations.” All human annotators were full-time employees of the company while performing this work.



\bibliography{paper}

\counterwithin{figure}{section}
\counterwithin{table}{section}

\clearpage
\appendix
\onecolumn

\section{Application of turn-level model on sentences}


\begin{figure*}[ht]
\centering
\begin{minipage}{1\linewidth}
\begin{algorithm}[H]

	\setcounter{AlgoLine}{0}
	\LinesNumbered 
	\setcounter{AlgoLine}{0}
	\BlankLine
	\LinesNumbered 
	\SetKwInOut{Input}{Input}
    \SetKwInOut{Output}{Output}
	
	\Input{Dialogue turns $\{T_j\}$ \\
	       Sentences $S_{jk} \in T_j$ \\
	       Universe of labels $\mathcal{L}$ \\
	       Model $M_\textnormal{turn}(T_j)$ for estimating section probability $P(L=L_i |T_j), L_i\in\mathcal{L}$
	       }
	\Output{Sentence level labels $L_{jk} \in \mathcal{L}$}
	\BlankLine	
	
    $L_{\textnormal{turn},j}   \leftarrow \{l \in \mathcal{L}: P(L=l |T_j) > \alpha_1\}$ \;
    
    $L_{\textnormal{filter},j}   \leftarrow \{l \in \mathcal{L}: P(L=l |T_j) > \alpha_2\}$ \;
	
	\ForEach{$S_{jk}\in T_j$}{
	    \If{$\textnormal{‘summarization’}\in L_{\textnormal{turn},j}$}{
	        $L_{jk} \leftarrow \textnormal{‘summarization’}$\;
	    }
	    \ElseIf{$P(L=\textnormal{“history taking”} |S_{jk}) \ge \alpha_3 $}{
	     $L_{jk} \leftarrow \textnormal{‘history taking’}$}
	    \ElseIf{$P(L=\textnormal{“education”} |S_{jk}) \ge \alpha_3 $}{
	     $L_{jk} \leftarrow \textnormal{‘education’}$}
	    \ElseIf{$P(L=\textnormal{“care plan”} |S_{jk}) \ge \alpha_3 $}{
	     $L_{jk} \leftarrow \textnormal{‘care plan’}$}
	    \Else{
	        $l_{\textnormal{candidate}} \leftarrow \argmax_c  P(L=l |S_{jk}), l \in L_{\textnormal{filter},j} $ \;
	        
	        $L_{jk} \leftarrow l_{\textnormal{candidate}} \textnormal{ if } P(L=l_{\textnormal{candidate}} |S_{jk})>\alpha_1 \textnormal{ else } \textnormal{“other”}$
	    }
	    }

	\Return $L_{jk}$
      
	\caption{Pseudocode for applying the turn-level model to create sentence-level labels }
   \label{algo:turn-to-sent}
\end{algorithm}

\end{minipage}
\end{figure*}

Where $\alpha_1=0.5$, $\alpha_2=0.1$, and $\alpha_3=0.9$. The values were determined by an informal human evaluation of the pseudolabeling performance.

\clearpage
\section{Example of Clustering outputs}
 \begin{figure*}[h!]
    \centering
    \includegraphics[width=\textwidth]{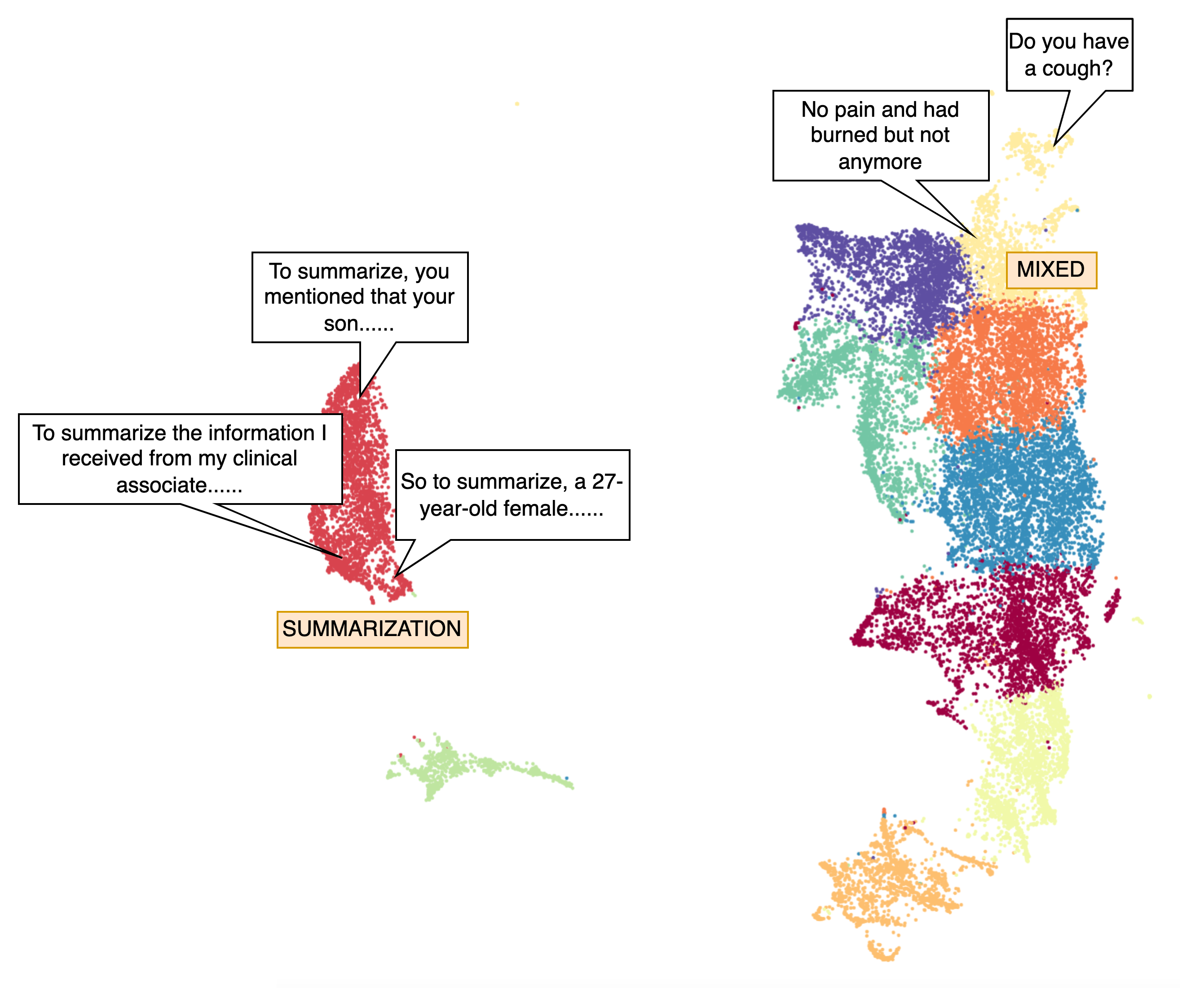}
    \caption{Clustering of Sentences Predicted as ``Summarization" after the First Round of Training }
    \label{fig:cluster}
\end{figure*}

\newpage
\section{Examples of Encounters with Color-coded Model-generated Predictions}

As our model is trained on predicting professionals' sentences, only the professionals' sentences are color-coded here.

\begin{figure}[!ht]
    \centering
    \includegraphics[width=0.2\linewidth,left]{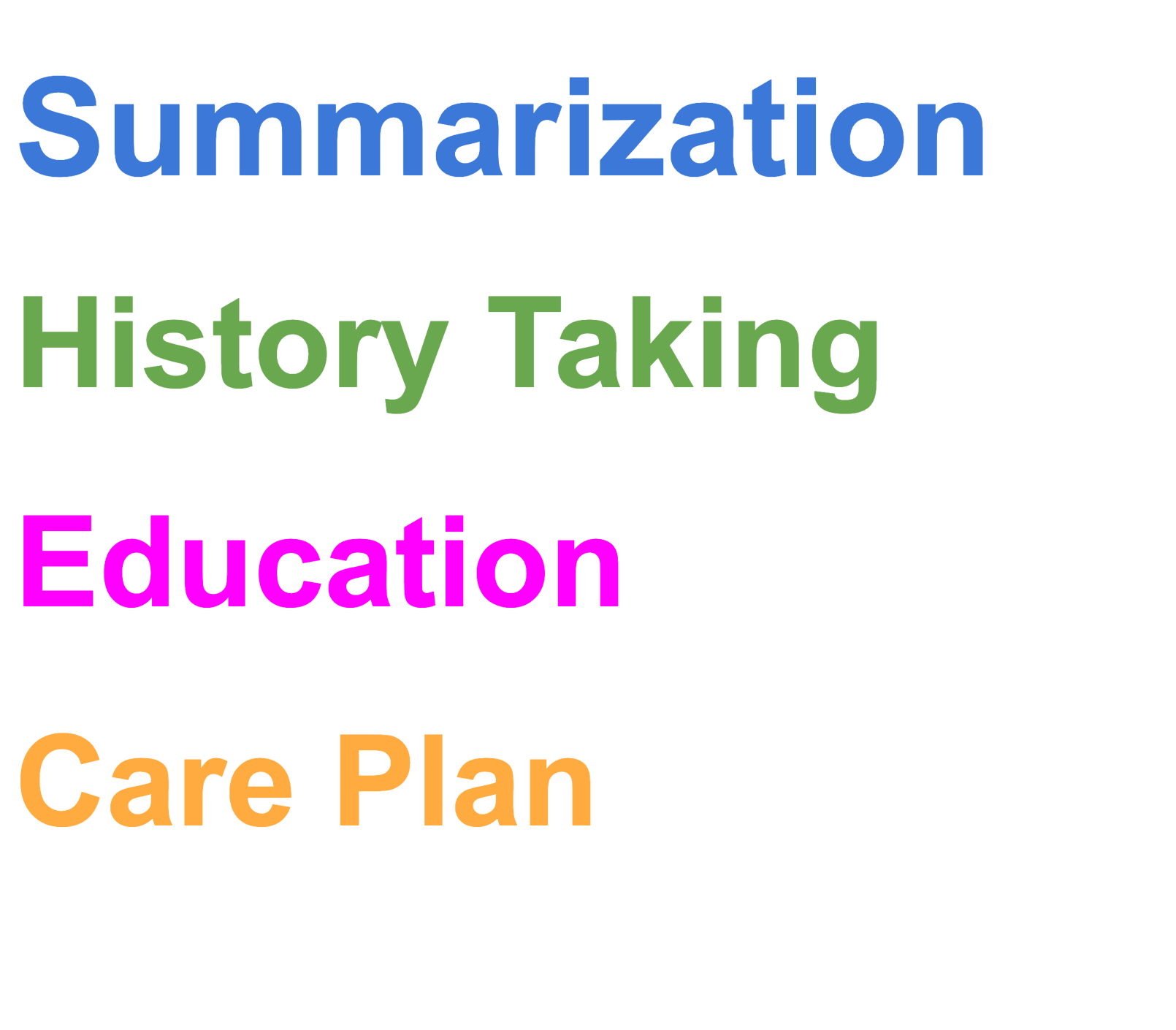}
    \label{fig:color_coding}
\end{figure}

\begin{figure}[!ht]
    \centering
    \includegraphics[width=1\linewidth]{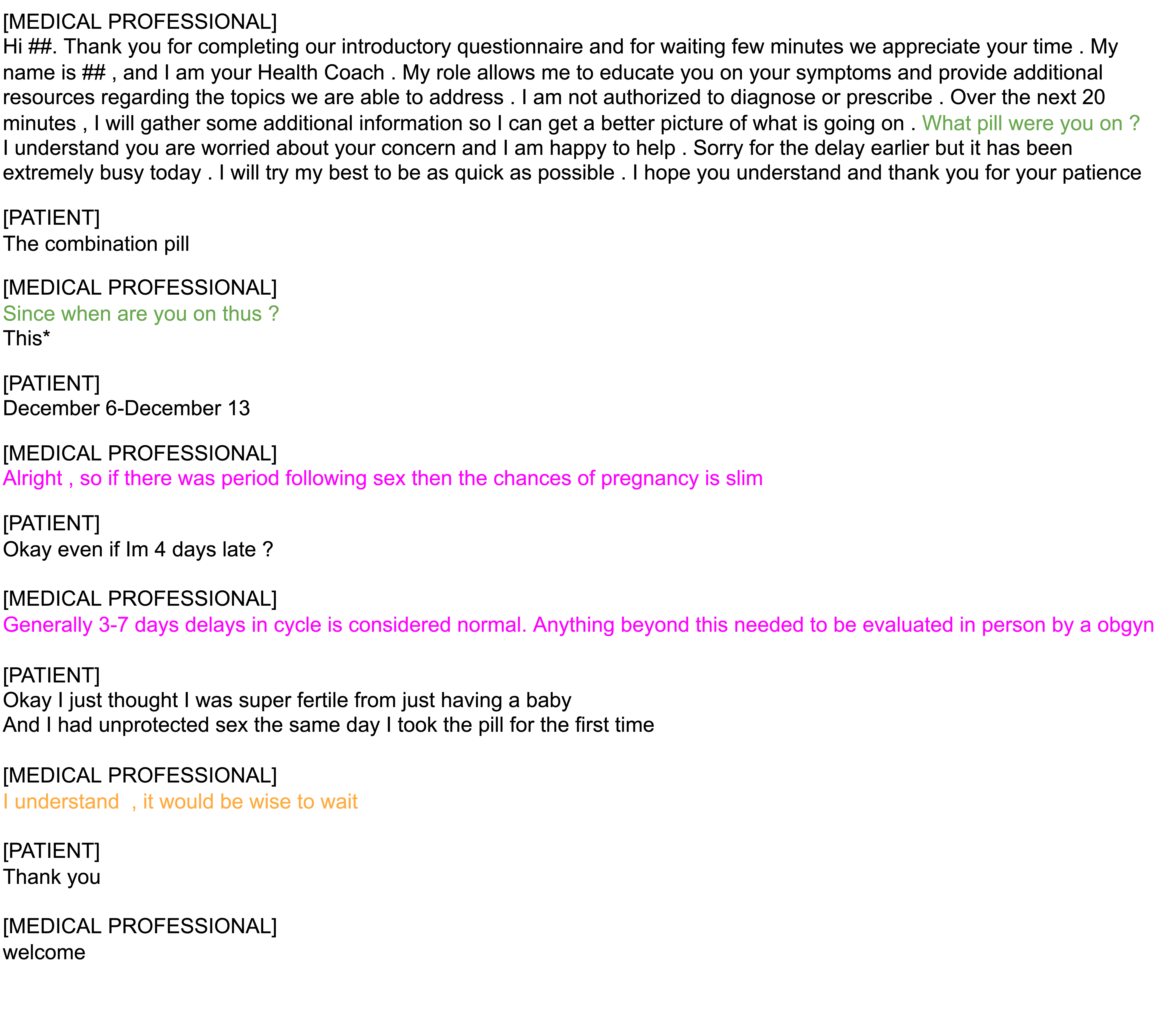}
    \caption{Sample Encounter 1}
    \label{fig:sample_conv_1}
\end{figure}

\begin{figure}[!ht]
    \centering
    \includegraphics[width=1\linewidth]{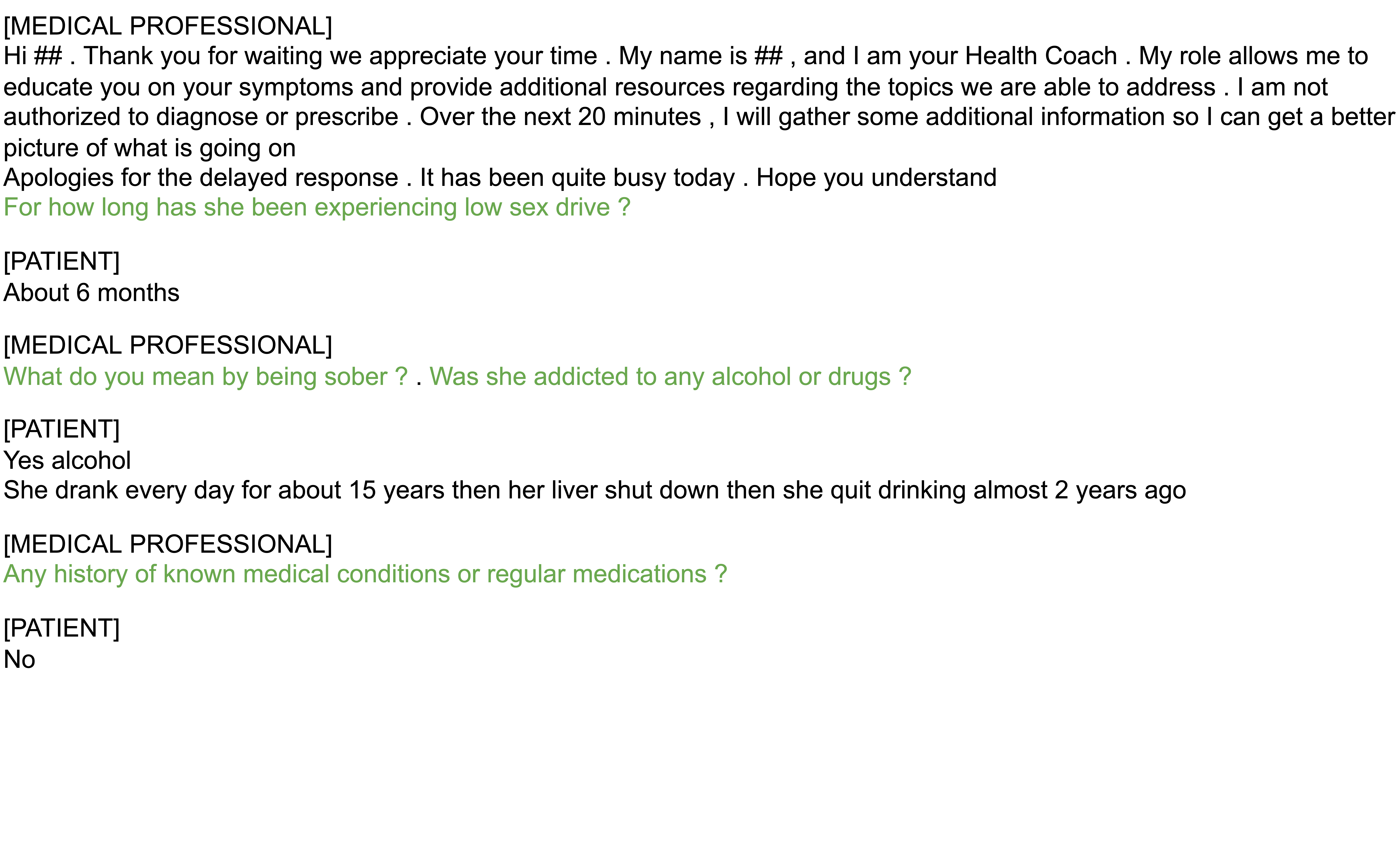}
    \caption{Sample Encounter 2}
    \label{fig:sample_conv_2}
\end{figure}

\begin{figure}[!ht]
    \centering
    \includegraphics[width=1\linewidth]{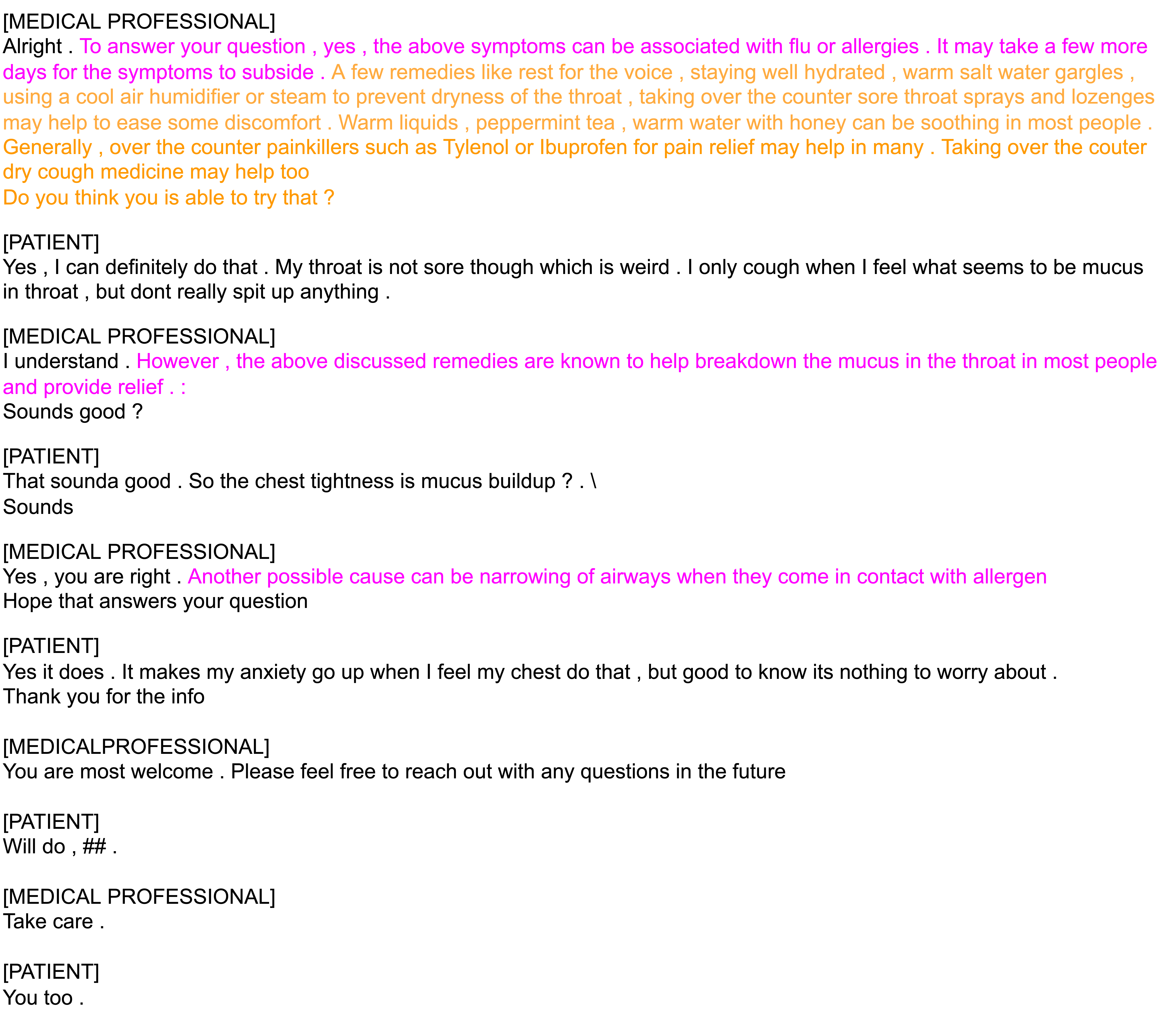}
    \caption{Sample Encounter 3}
    \label{fig:sample_conv_3}
\end{figure}

\begin{figure}[!ht]
    \centering
    \includegraphics[width=1\linewidth]{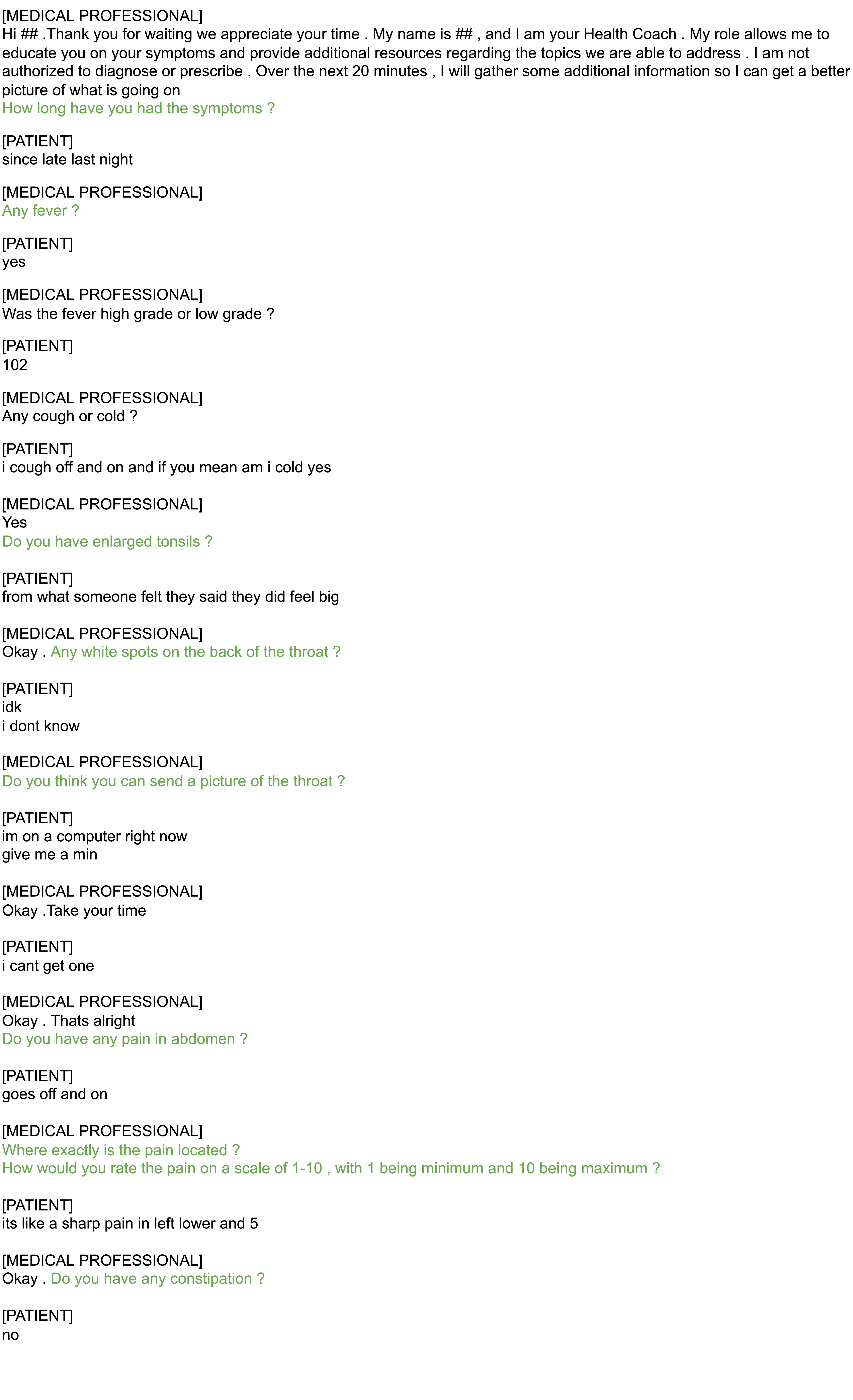}
    \label{fig:sample_conv_4_1}
\end{figure}

\begin{figure}[!ht]
    \centering
    \includegraphics[width=1\linewidth]{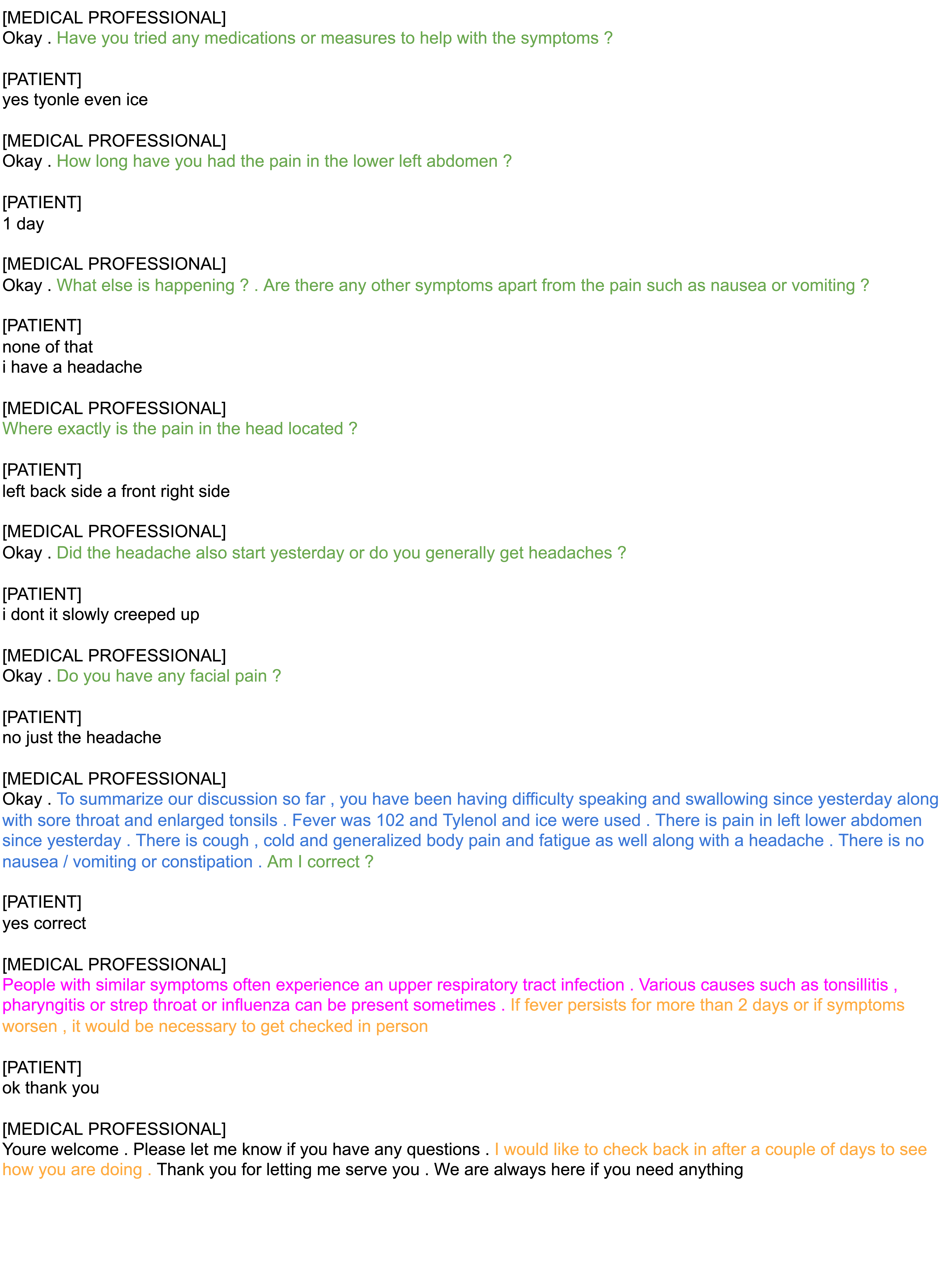}
    \caption{Sample Encounter 4}
    \label{fig:sample_conv_4_2}
\end{figure}


\end{document}